\documentclass[letterpaper]{article}
\usepackage[draft]{aaai25}
\usepackage{times}
\usepackage{helvet}
\usepackage{courier}
\usepackage[hyphens]{url}
\usepackage{graphicx}
\usepackage{natbib}
\usepackage{caption}
\usepackage{algorithm}
\usepackage{algorithmic}
\usepackage{amsmath}
\usepackage{multirow}
\usepackage{booktabs}
\usepackage{amsfonts}
\usepackage{newfloat}
\usepackage{listings}
\DeclareCaptionStyle{ruled}{labelfont=normalfont,labelsep=colon,strut=off}
\floatstyle{ruled}
\newfloat{listing}{tb}{lst}{}
\floatname{listing}{Listing}
\title{VPO: Leveraging the Number of Votes in Preference Optimization}
\author{
    Jae Hyeon Cho\textsuperscript{\rm 1},
    Minkyung Park\textsuperscript{\rm 1},
    Byung-Jun Lee\textsuperscript{\rm 1}
}
\affiliations{
    \textsuperscript{\rm 1}Department of Artifcial Intelligence\\
    Korea University, Seoul, Republic of Korea\\
    \{bonin147, swwwjkl538, byungjunlee\}@korea.ac.kr
}

\DeclareMathOperator*{\argmin}{argmin}
\begin{document}

\maketitle

\begin{abstract}
Direct Preference Optimization (DPO) trains a language model using human preference data, bypassing the explicit reward modeling phase of Reinforcement Learning from Human Feedback (RLHF). By iterating over sentence pairs in a preference dataset, DPO enhances generation quality by increasing the likelihood of producing preferred sentences over less favored ones. Preference datasets are typically created by selecting preferred sentences through a voting process involving multiple individuals, as opinions can vary due to the subjective nature of human preferences. While the number of votes offers insight into whether a sentence pair is clearly preferable or controversial, current methods do not fully leverage this information. In this paper, we introduce a technique that leverages user voting data to better align with diverse subjective preferences. We employ the Bayesian Minimum Mean Square Error (Bayesian MMSE) estimator to model the probability that one generation is preferable to another. Using this estimated probability as a target, we develop the Vote-based Preference Optimization (VPO) framework, which incorporates the number of votes on both sides to distinguish between controversial and obvious generation pairs. We show that previous algorithms, such as DPO and Identity Preference Optimization (IPO), can be extended using the proposed framework, termed VDPO and VIPO. Our experiments demonstrate that these proposed algorithms outperform various existing methods, including their base algorithms.
\end{abstract}

\begin{figure}[t!]
  \centerline{\includegraphics[width=\columnwidth]{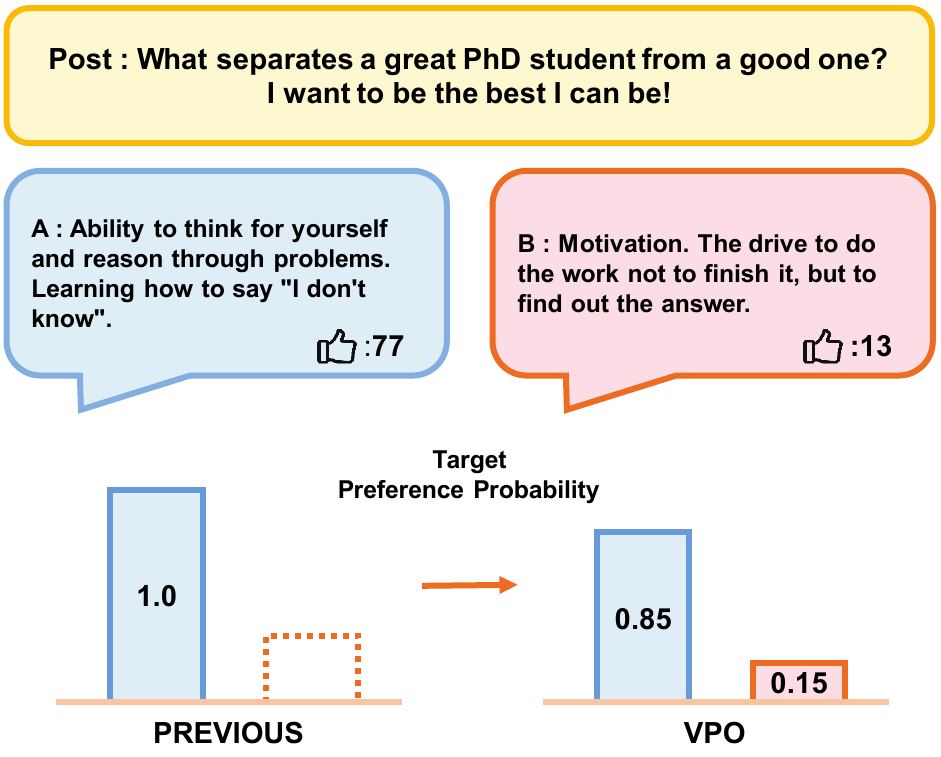}}
  \caption{While previous methods trained models to generate responses based on majority preference (e.g., A), human preferences are subjective, making responses like B also desirable. Our proposed framework, VPO, utilizes additional information to capture a more nuanced understanding of these preferences.}
  \label{fig:introduction}
\end{figure}

\section{Introduction}
In general-domain applications of language models (LM), the model should be aligned with human values, such as helpfulness, honesty, and harmlessness. 
Pre-training and supervised fine-tuning (SFT) enable the development of models with notable capabilities across a wide range of natural language processing (NLP) tasks \citep{wei2021finetuned, wang2023far}. However, additional training using pairwise preference data is often employed to further align the model with human values.

Preference alignment methods, such as reinforcement learning from human feedback (RLHF, \citeauthor{stiennon2020learning} 2020; \citeauthor{ouyang2022training} 2022) and direct preference optimization (DPO, \citeauthor{rafailov2023direct} 2023), have shown significant successes in enhancing the human usability of language models. Consequently, these preference optimization processes are now considered essential in the development of state-of-the-art large LMs \citep{achiam2023gpt,team2023gemini}.

Given pairwise preference data with labels indicating which response is preferred, RLHF trains a reward model to align with these preferences, enabling the evaluation of a language model's outputs. Subsequently, the language model is trained using a reinforcement learning algorithm to maximize the expected reward of its generated responses. In contrast, DPO provides an alternative approach by directly adjusting the generation probabilities of the language model based on preference labels. This method eliminates the need for a separate reward modeling phase, thereby reducing computational costs.

However, we note that the current labels in pairwise preference datasets may provide limited information in these processes. Human preference is inherently subjective, and not all pairwise generations can be easily classified as simply better or worse, as judgments vary among individuals. As a result, voting or scoring processes are often utilized to gather preference data, yet this additional information has largely been overlooked in previous studies on preference alignment.

In this paper, we introduce a simple yet effective method to better utilize the rich side information inherent in human preference datasets.
Our approach models the underlying target preference probability using the Bayesian Minimum Mean Square Error (MMSE) estimator, enabling the model to distinguish between learning from clear-cut examples (those with a high vote/score gap) and contentious cases (those with a low vote/score gap).
We term this framework as Vote-based Preference Optimization (VPO), and extend established algorithms such as DPO and Identity Preference Optimization (IPO, \citeauthor{azar2024general} 2024) into VDPO and VIPO, respectively, demonstrating the broad applicability of our approach.

In the experiments, we empirically demonstrate the following:
\begin{itemize}
    \item VDPO and VIPO outperform existing algorithms, achieving improved generation quality and training stability.
    \item Our framework is also adaptable to utilize AI feedback in scenarios where costly human voting information is unavailable.
\end{itemize}

\section{Preliminaries}
In this section, we outline the standard procedures for training a general-purpose language model prior to aligning it with human values. The process begins with the following two steps:
\paragraph{Pretraining} To provide the model with general capabilities, it is trained on a large corpus using next token prediction, commonly referred to as teacher forcing.
\paragraph{Supervised finetuning} Following pretraining, supervised finetuning is performed to equip the model with specific abilities required for the target domain tasks. 
During this phase, the model is trained on a dataset specifically curated for the intended tasks. 
We refer to the model after this step as $\pi_{\text{ref}}$ henceforth.

\subsection{RLHF: Alignment via Reward Modeling}
The standard RLHF process consists of two steps.
\paragraph{Reward model training} The reward model is trained using human preference data to align its judgments with human values. The human preference dataset is composed of the triplet $\mathcal{D} = \{x,y_{1},y_{2}\}$, where $x$ is the context, and $y_1$ and $y_2$ are response pairs given the context.
We define the events:
\begin{align*}
    Y_1&=\text{$y_1$ is favored over $y_2$},\quad Y_2=\text{$y_2$ is favored over $y_1$}.
\end{align*}
The probability of these events is modeled using a Bradley-Terry model, which is defined as follows:
\begin{equation*}
    \hat{p}_r(Y_{1}|x):=\frac{\exp(r(x,y_1))}{\exp(r(x,y_1))+\exp(r(x,y_2))}
\end{equation*}
The reward model is then optimized by maximizing the log-likelihood of the human preference dataset,
\begin{equation}
\label{eq:reward loss}
    \max_r \mathbb{E}_{(x,y_1,y_2)\sim \mathcal{D}}\big[\log \hat{p}_r(Y_1|x)\big],
\end{equation}
assuming, without loss of generality, that $y_1$ is the preferred response.
\paragraph{RL finetuning} 
After training a reward model, a regularized RL algorithm is used to maximize the expected reward while ensuring the model does not deviate significantly from the initial model $\pi_{\text{ref}}$:
\begin{equation*}
\max_{\theta} \mathbb{E}_{x \sim \mathcal{D}, y \sim \pi_\theta(y|x)} \big[ r(x,y) -\beta D_{\text{KL}}(\pi_\theta(\cdot|x) \| \pi_{\text{ref}}(\cdot|x)) \big].
\end{equation*}
This approach ensures that the updated model achieves high reward, meaning strong alignment with human preferences, while preserving the general language capabilities of the reference model.
\subsection{DPO: Alignment without Reward Modeling}
\label{sec:DPO}
\paragraph{Direct preference optimization}
Training an additional reward model, along with using reinforcement learning to fine-tune a model, involves numerous complex engineering challenges. DPO provides an alternative approach by directly training the language model on the preference dataset by substituting reward model with its closed-form solution.

Assuming $y_1$ is preferred over $y_2$, the DPO objective is defined as follows:
\begin{align}
\label{eq:DPO Loss}
    \max_\theta~\mathbb{E}_{\mathcal{D}}&\left[\log \sigma (r(x,y_1)- r(x,y_2))\right],\\
    \text{where}\quad r(x,y)&=\beta \log \dfrac{\pi_{\theta}(y|x)}{\pi_{\text{ref}}(y|x)} + \beta \log Z(x), \nonumber\\
    \text{and} \quad
    Z(x) &= \sum_{y} \pi_{\text{ref}}(y|x) \exp\left(\dfrac{1}{\beta}r(x,y)\right).\nonumber
\end{align}
DPO simplifies the training process by directly leveraging the dataset, thereby enhancing both stability and efficiency.

\paragraph{Reward divergence}
A significant drawback of DPO, as highlighted by \citet{azar2024general}, is that its objective is theoretically prone to divergence. When considering a single data point $(x, y_1, y_2)$, the DPO objective continually drives an increase in the margin $r(x,y_1)-r(x,y_2)$ without bound. In practice, this results in an inflated scale of the implicit reward function, which undermines the regularization towards $\pi_{\text{ref}}$. This is one reason why DPO often becomes unstable when trained over multiple epochs and requires early stopping.

To address this issue, Identity Preference Optimization (IPO, \citeauthor{azar2024general} 2024) and conservative DPO (cDPO, \citeauthor{mitchell2023cdpo} 2023) have been proposed, both of which stabilize training by adjusting the objective.

\iffalse
 proposed the Identity Preference Optimization (IPO) algorithm, which aims to stabilize the training process.
Furthermore,  proposed Conservative DPO (cDPO) to address reward divergence:
\begin{equation}
    \mathcal{L}_{\textrm{cDPO}}(\pi_\theta;\pi_{\text{ref}})
    =
    (1-\epsilon)\mathcal{L}_{\textrm{DPO}}(y_1,y_2)
    +\epsilon\mathcal{L}_{\textrm{DPO}}(y_2,y_1)
\end{equation}
where, $\epsilon$ represents the noise in preferences.
This objective can be understood as $\mathcal{L}_{\textrm{DPO}}$ with a $\epsilon$-label smoothing applied.
\fi

\begin{table*}[t!]
\centering
  \begin{tabular}{p{0.04\linewidth}p{0.90\linewidth}}
  \toprule[1.5pt]
  \centering \multirow{3}{*}{Post} 
  & Anybody else almost always reduce the sugar in recipes? I just made a cheesecake with half the sugar and it was delicious.  I do this a lot with cakes, tarts and muffins (what I usually make) and have never had any problems, so I do wonder why recipes contain such a high amount of sugar...\\
  \bottomrule[1pt]
  \multicolumn{2}{c}{}\\
  \toprule[1pt]
  Vote & \multicolumn{1}{c}{Comment}\\
  \midrule
  \centering\multirow{2}{*}{101}    & $y_1:$ **It's about balance.** Your cheesecake has a lot of rich ingredients, like 8 ounces of cream cheese, 1/2 cup  sour cream, 5 eggs, and a ton of butter in the crust. The sugar balances the richness from these ingredients ...\\ \midrule
  \centering 15     & $y_2:$ I always cut the sugar in half. I want to taste everything in my dessert, not just sugar. \\ \midrule
  \centering 14     & $y_3:$ I almost always cut it by 1/4 to a 1/2. I like to taste flavors not just sugar and my teeth don’t feel as gritty either. \\ \midrule
  \centering\multirow{2}{*}{9}     & $y_4:$ I live in Brazil and the overall palate and traditional recipes here are always too sweet for me. I tend to dial down everything I make for myself. But, Im a pastry chef, and things I do for my job tend to be a little sweeter... \\  
  \bottomrule[1.5pt]
  \end{tabular}
  \caption{Example from the SHP dataset illustrating a post and four comments with different vote counts. Conventional approaches consider the relationship between $(y_2, y_3)$ the same as between $(y_1, y_4)$, which is undesirable. This example demonstrates the limitations of reducing vote information to a simple binary indication of more or less preferred.}
  \label{tab:illustrative_example}
\end{table*}

\section{Related Works}
\paragraph{Alignment without reward modeling}
Since the introduction of DPO, several studies have focused on improving the efficiency of preference alignment algorithms. As briefly introduced above, \citet{azar2024general} mathematically analyzed the issue of diverging rewards in DPO and proposed IPO as as a potential solution.
\citet{ethayarajh2024kto} introduced Kahneman-Tversky Optimization (KTO), which utilizes the Kahneman-Tversky human utility function to better align with human values.
\citet{hong2024orpo} presented the Odds Ratio Preference Optimization (ORPO), a reference model-free approach that eliminates the dependency on a baseline model, thereby simplifying the optimization process. 

Although various improvements to DPO are being explored, they still share the limitation of not fully utilizing side information beyond the binary indication of more or less preferred. In this paper, we propose enhancing existing algorithms by incorporating additional side data. The improvements we suggest are orthogonal to the advancements made by these existing methods and can be seamlessly integrated into all of these approaches.

\paragraph{Noise in preference labels}
Several studies have examined the potential for preference labels to be noisy due to human subjectivity. While the primary objective of cDPO~\citep{mitchell2023cdpo} was to address the issue of reward divergence, the algorithm was formulated with the assumption that preference labels may contain noise. To further enhance the robustness of learning in noisy environments, \citet{chowdhury2024provably} developed robust DPO (rDPO), which is specifically designed to minimize the impact of noise in preference labels.

As we will demonstrate, our VPO framework can also be interpreted as modeling the level of noise in preference labels using side information. In cDPO and rDPO, this noise level is assumed to be constant and is tuned as a hyperparameter. In contrast, our approach offers a straightforward and intuitive method for estimating noise levels in the preference dataset, building on a similar framework.

\section{Method}
\label{sec:Method}
In standard protocols for constructing human preference datasets, each generation pair is typically evaluated multiple times by different evaluators to account for the variability in their judgments.
Although the number of votes from these evaluators is usually recorded during dataset creation, this information has often been underutilized in previous methodologies. Below, we provide a detailed illustrative example to emphasize this point.

\paragraph{Illustrative example} Table~\ref{tab:illustrative_example} presents an example of the raw data format (post and comments) from the Stanford Human Preference (SHP) dataset.
For clarity, long contexts and responses have been truncated, with the full example available in the Appendix \ref{sec:Full Representation of SHP Dataset Examples}.

Using the conventional approach to align a language model, we lose valuable side information, leading the model to be trained to prefer $y_1$ over $y_2$, $y_2$ over $y_3$, $y_3$ over $y_4$, all with the same margins between them. However, a human evaluator would likely judge that $y_1$ should be preferred over the others by a significant margin, while the other three responses---$y_2$, $y_3$, and $y_4$--- are of rougly equal quality. 

To this end, we propose modeling the \textit{target preference probability}: $p(Y_1|x, v_1, v_2)$ and $p(Y_2|x, v_1, v_2)$, where $v_1$ and $v_2$ represent the number of votes for $y_1$ and $y_2$, respectively. In previous approaches, it is typically assumed that $y_1$ is the preferred response, assigning $p(Y_1|x)=1$ and $p(Y_2|x)=0$. Instead, we employ a Bayesian approach to model the target preference probability, taking into account the number of votes $v_1, v_2$ collected during dataset construction. This approach allows for a nuanced interpretation of vote counts, enabling the distinction between different vote distributions, ranging from controversial response pairs to more obvious ones.

\subsection{Modelling Targets with Bayesian MMSE}
\label{sec:Bayesian MMSE}
To better align with the human preference through the finite number of assessments in the preference dataset, we adopt a Bayesian approach. We begin by defining the prior distribution of the target preference probability when no voting information is observed:
\begin{equation*}
    p(Y_{1}|x)\sim \text{Beta}(c,c)\quad \text{where $c>0$ is a hyperparameter.}
\end{equation*}
Since the response pair can appear in any order, the distribution should be order-invariant. To ensure this, the parameters of the Beta distribution are set equal, both taking the value $c$.
The hyperparameter $c$ controls the influence of the number of votes on the posterior distribution: larger values of $c$ reduce the impact of voting, while smaller values amplify it.

Let $p(Y_{1}|x)$ be denoted as $\theta$.
The number of votes for each option is represented by the non-negative integers $v_1$ and $v_2$.
We model the likelihood function for human preferences using a binomial distribution, which simplifies the computation of the posterior due to its conjugate properties:
\begin{align*}
    p(v_1,v_2|\theta)&\propto\theta^{v_1}(1-\theta)^{v_2},\\
    p(\theta|v_1,v_2)&=\text{Beta}(v_1+c,v_2+c).
\end{align*}

A straightforward approach to utilizing the posterior distribution $p(\theta \mid v_1, v_2)$ would be to sample $\theta$ each time the language model is updated based on a response pair. However, to better stabilize the training of a large model, we employ the Bayesian Minimum Mean Square Error (MMSE) estimator, which involves simply taking the mean:
\begin{equation}
    \hat{\theta}_{\text{MMSE}}(v_1,v_2)
    =\mathbb{E}[\theta|v_1,v_2]=\frac{v_1+c}{v_1+v_2+2c}, \label{mmse}
\end{equation}
where the name derives from its property:
\newtheorem{theorem}{Theorem}
\begin{theorem}\cite{pishro-nik2014}
    Bayesian MMSE estimator is solution to the following:
\begin{align*}
    \hat{\theta}_{\text{MMSE}}(v_1,v_2)
    &=\argmin_{\hat{\theta}}\int(\hat{\theta}-\theta)^2p(\theta|v_1,v_2)d\theta.
\end{align*}
\end{theorem}

\begin{table}
  \centering
  \begin{tabular}{cc c}
  \toprule[1.5pt]
    votes & $\hat{\theta}_{\text{MMSE}}(v_1,v_2)$ & \textbf{$p(Y_1|x)$} \\
    \midrule
    101:9     & 0.91     & 1     \\
    15:14    & 0.516     & 1     \\
    14:9       & 0.6     & 1     \\
  \bottomrule[1.5pt]
  \end{tabular}
  \caption{The estimated probability based on Bayesian MMSE estimator for different vote count with $c=1$, compared to the target preference probability we had with DPO.}
  \label{tab:mmse}
\end{table}

Using Bayesian MMSE estimator allows us to convey the implication of various number of votes of response pairs without introducing additional stochasticity to the training.

\paragraph{Illustrative example} Table~\ref{tab:mmse} provides examples of the Bayesian MMSE estimator with the hyperparameter $c=1$.
\begin{itemize}
    \item For clear-cut response pairs such as $101:9$, the estimator indicates a strong preference of $0.91$.
    \item For controversial pairs like $15:14$, the estimator shows a much weaker preference at $0.52$.
    \item In the case of the pair $14:9$, although the ratio suggest a significant preference for the first response, the estimator provides a moderate preference of $0.6$, acknowledging that the vote count does not provide enough evidence.
\end{itemize}
  These examples demonstrate how the Bayesian MMSE estimator enables the language model to learn differently from various response pairs, taking into account the number of votes to better align with subjective human preferences.

\subsection{Vote-based Preference Optimization (VPO)}
\label{sec:VPO}
Adopting the Bayesian MMSE estimator as the target preference probability, $p(Y_1|x,v_1,v_2)=\hat{\theta}_{\text{MMSE}}(v_1,v_2)$,
creates a versatile framework that can be generalized to extend various preference optimization algorithms.
We refer to these collection of extended algorithms as the \textbf{Vote-based Preference Optimization (VPO)} framework, which enables a more nuanced understanding of subjective human preferences. 

\paragraph{Cross entropy with generalized targets} In previous approaches, including RLHF and DPO, the (implicit) reward function is trained by maximizing the log-likelihood, as shown in Eq.~\eqref{eq:reward loss}. This can be interpreted as assuming $p(Y_{1}|x)=1$ as the target and using a cross entropy objective. By adopting the generalized target probability $p(Y_1|x,v_1,v_2)$ from VPO, we now obtain a generalized reward loss function:
\begin{equation}
\label{eq:Binary Cross Entropy Loss}
\max_r\mathbb{E}_{\mathcal{D}}\left[\sum_{i=1}^{2}p(Y_i|x,v_1,v_2)\log \hat{p}_r(Y_i|x)\right].
\end{equation}
This objective functions as an adaptive label smoothing mechanism, ensuring that the reward function learns to have a large reward margin for substantial vote gaps and a smaller reward margin for narrower vote gaps.
\paragraph{Vote-based Direct Preference Optimization (VDPO)}
To implement our approach within DPO, we generalize Eq.~\eqref{eq:DPO Loss} using the target preference probability from VPO:
\begin{align}
\max_\theta~\mathbb{E}_{\mathcal{D}}\big[&p(Y_1|x,v_1,v_2)\log \sigma \big(r(x,y_1)- r(x,y_2)\big) \nonumber \\
    +&p(Y_2|x,v_1,v_2)\log \sigma \big(r(x,y_2)- r(x,y_1)\big)\big],
\end{align}
where $r(\cdot)$ is defined as in Eq.~\eqref{eq:DPO Loss}.
In addition to differentiating response pairs with varying vote ratios, as discussed in \citet{mitchell2023cdpo}, including both the more preferred and the less preferred responses contributes to more stable training by addressing the issue of reward divergence.

\paragraph{Vote-based Identity Preference Optimization (VIPO)}
While IPO~\citep{azar2024general} was introduced to address reward divergence, it can still benefit from distinguishing pairs with varying vote ratios by incorporating VPO. Its objective is:
\begin{equation}
\label{eq:IPO}
\min_\theta \mathbb{E}_{\mathcal{D}}\left[\left(r(x,y_1)-r(x,y_2)-\frac{1}{2\beta}\right)^2\right],
\end{equation}
which tries to fix the reward margin to be $\frac{1}{2\beta}$. This objective is derived from:
\begin{align*}
\min_\theta \mathbb{E}_{\mathcal{D}}\Big[&\left(r(x,y_1)-r(x,y_2)-\beta^{-1}p(Y_1|x)\right)^2\\
+&\left(r(x,y_2)-r(x,y_1)-\beta^{-1}p(Y_2|x)\right)^2\Big],
\end{align*}
with $p(Y_1|x)=1$ and $p(Y_2|x)=0$. Adopting VPO by substituting $p(Y_1|x,v_1,v_2)$ in, up to a constant we get:
\begin{equation*}
    \min_\theta\mathbb{E}_{\mathcal{D}}\left[\left(r(x,y_1)-r(x,y_2)-\frac{2p(Y_1|x,v_1,v_2)-1}{2\beta}\right)^2\right]
\end{equation*}
By leveraging vote-based information, VIPO adjusts the reward margin to be proportional to the strength of human preference, up to a maximum of $\frac{1}{2\beta}$.

\begin{table*}[t!]
  \centering
  \begin{tabular}{cccccc}
  \toprule[1.5pt]
    Pretrained & \multirow{2}{*}{Algorithm} & \multicolumn{2}{c}{In-domain} & \multicolumn{2}{c}{Alpaca Farm}\\
    Model & & Win rate & LC Win rate & Win rate & LC Win rate \\
    \midrule
    \multirow{9}{*}{Pythia 2.8B}
    & DPO & $\text{52.88}_{(\pm\text{2.03})}$&$\text{51.84}_{(\pm\text{0.15})}$ & $\text{55.92}_{(\pm\text{1.46})}$&$\text{55.88}_{(\pm\text{0.10})}$ \\
    & IPO & $\text{50.89}_{(\pm\text{2.08})}$&$\text{49.59}_{(\pm\text{0.10})}$ & $\text{56.35}_{(\pm\text{1.47})}$&$\text{55.99}_{(\pm\text{0.09})}$ \\
    & KTO & $\text{47.03}_{(\pm\text{2.07})}$&$\text{46.76}_{(\pm\text{0.16})}$ & $\text{51.05}_{(\pm\text{1.49})}$&$\text{50.70}_{(\pm\text{0.11})}$ \\
    & cDPO (0.1) & $\text{49.50}_{(\pm\text{2.10})}$&$\text{49.25}_{(\pm\text{0.19})}$ & $\text{51.63}_{(\pm\text{1.50})}$&$\text{51.28}_{(\pm\text{0.16})}$ \\
    & cDPO (0.3) & $\text{50.63}_{(\pm\text{2.07})}$&$\text{50.65}_{(\pm\text{0.20})}$ & $\text{49.61}_{(\pm\text{1.49})}$&$\text{49.83}_{(\pm\text{0.14})}$ \\
    & rDPO (0.1) & $\text{50.43}_{(\pm\text{2.06})}$&$\text{49.25}_{(\pm\text{0.19})}$ & $\text{51.13}_{(\pm\text{1.48})}$&$\text{49.92}_{(\pm\text{0.06})}$ \\
    & rDPO (0.3) & $\text{50.15}_{(\pm\text{2.04})}$&$\text{49.25}_{(\pm\text{0.27})}$ & $\text{49.92}_{(\pm\text{1.48})}$&$\text{49.83}_{(\pm\text{0.14})}$ \\
    & VDPO (ours) & $\text{53.37}_{(\pm\text{2.08})}$&$\text{52.08}_{(\pm\text{0.21})}$ & $\textbf{57.05}_{(\pm\text{1.48})}$&$\textbf{56.70}_{(\pm\text{0.13})}$ \\
    & VIPO (ours) & $\textbf{54.75}_{(\pm\text{2.06})}$&$\textbf{53.14}_{(\pm\text{0.17})}$ & $\text{56.49}_{(\pm\text{1.48})}$&$\text{56.43}_{(\pm\text{0.12})}$ \\
    \midrule
    \multirow{9}{*}{LLaMA 7B}
    & DPO & $\text{42.10}_{(\pm\text{2.22})}$&$\text{28.04}_{(\pm\text{0.21})}$ & $\text{32.66}_{(\pm\text{1.45})}$&$\text{24.19}_{(\pm\text{0.13})}$ \\
    & IPO & $\text{48.84}_{(\pm\text{2.30})}$&$\text{42.53}_{(\pm\text{0.34})}$ & $\text{51.88}_{(\pm\text{1.57})}$&$\text{48.53}_{(\pm\text{0.19})}$ \\
    & KTO & $\text{45.52}_{(\pm\text{2.21})}$&$\text{33.57}_{(\pm\text{0.43})}$ & $\text{37.27}_{(\pm\text{1.49})}$&$\text{30.33}_{(\pm\text{0.19})}$ \\
    & cDPO (0.1) & $\text{42.32}_{(\pm\text{2.22})}$&$\text{28.88}_{(\pm\text{0.18})}$ & $\text{34.97}_{(\pm\text{1.47})}$&$\text{24.95}_{(\pm\text{0.14})}$ \\
    & cDPO (0.3) & $\text{48.36}_{(\pm\text{2.24})}$&$\text{37.95}_{(\pm\text{0.28})}$ & $\text{52.12}_{(\pm\text{1.57})}$&$\text{45.21}_{(\pm\text{0.12})}$ \\
    & rDPO (0.1) & $\text{36.51}_{(\pm\text{2.17})}$&$\text{24.31}_{(\pm\text{0.41})}$ & $\text{28.14}_{(\pm\text{1.38})}$&$\text{21.38}_{(\pm\text{0.18})}$ \\
    & rDPO (0.3) & $\text{39.56}_{(\pm\text{2.17})}$&$\text{25.44}_{(\pm\text{0.27})}$ & $\text{26.55}_{(\pm\text{1.36})}$&$\text{18.09}_{(\pm\text{0.13})}$ \\
    & VDPO (ours) & $\textbf{51.81}_{(\pm\text{2.23})}$&$\text{41.35}_{(\pm\text{0.28})}$ & $\textbf{55.42}_{(\pm\text{1.56})}$&$\text{49.63}_{(\pm\text{0.14})}$ \\
    & VIPO (ours) & $\text{49.62}_{(\pm\text{2.29})}$&$\textbf{47.62}_{(\pm\text{0.20})}$ & $\text{51.69}_{(\pm\text{1.54})}$&$\textbf{49.74}_{(\pm\text{0.18})}$ \\
    \bottomrule[1.5pt]
  \end{tabular}
  \caption{Results on the SHP dataset evaluated using AlpacaEval. The table shows the win rates (\%) of various models compared to the SFT model, along with their standard deviations. LC denotes length-controlled. Our VDPO and VIPO models consistently outperform other models, showing improvements over DPO and IPO across all evaluated metrics.}
  \label{tab:evalSHP}
\end{table*}

\begin{table*}[t!]
  \centering
  \begin{tabular}{cccccc}
  \toprule[1.5pt]
    Pretrained & \multirow{2}{*}{Algorithm} & \multicolumn{2}{c}{In-domain} & \multicolumn{2}{c}{Alpaca Farm}\\
    Model & & Win rate & LC Win rate & Win rate & LC Win rate \\
    \midrule
    \multirow{4}{*}{Pythia 2.8B}
    & DPO & $\text{50.10}_{(\pm\text{1.87})}$&$\text{52.13}_{(\pm\text{0.16})}$ & $\text{53.92}_{(\pm\text{1.49})}$&$\text{58.12}_{(\pm\text{0.09})}$ \\
    & IPO & $\text{53.74}_{(\pm\text{1.84})}$&$\text{53.62}_{(\pm\text{0.25})}$ & $\text{50.94}_{(\pm\text{1.48})}$&$\text{50.58}_{(\pm\text{0.11})}$ \\
    & VDPO (ours) & $\text{57.40}_{(\pm\text{1.85})}$&$\text{56.82}_{(\pm\text{0.19})}$ & $\text{56.90}_{(\pm\text{1.48})}$&$\text{56.55}_{(\pm\text{0.08})}$ \\
    & VIPO (ours) & $\text{54.32}_{(\pm\text{1.84})}$&$\text{54.10}_{(\pm\text{0.16})}$ & $\text{51.93}_{(\pm\text{1.48})}$&$\text{51.58}_{(\pm\text{0.12})}$ \\
    % \midrule
    \bottomrule[1.5pt]
  \end{tabular}
  \caption{Results on the UFB dataset evaluated using AlpacaEval. The table shows the win rates (\%) of various models compared to the SFT model, along with their standard deviations. LC denotes length-controlled.}
  \label{tab:evalUFB}
\end{table*}

\section{Experiments}
In this section, we outline the experimental settings used to evaluate the performance of the proposed VPO framework.
Additional details about the experiments are available in Appendix~\ref{sec:Experimental Details}.

\subsection{Training Details}
\paragraph{Data} Our experiments utilize two widely recognized binary human preference datasets: the Stanford Human Preferences dataset (SHP, \citeauthor{pmlr-v162-ethayarajh22a} 2022) and the UltraFeedback Binarized dataset (UFB, \citeauthor{cui2023ultrafeedback} 2023). 
\begin{itemize}
    \item The SHP dataset consists of Reddit data, where the voting score is calculated by subtracting the number of negative votes from the number of positive votes and then adding one to the result. We use the voting scores directly as $v_1$ and $v_2$ for computing the target preference probability $p(Y_i|x,v_1, v_2)$.
\item In contrast, the UFB dataset employs GPT-4 for score annotation, with scores ranging from 1 to 10. To account for the different scaling of this scoring mechanism compared to the number of votes in human-annotated datasets, we exponentiated the scores before computing the target preference probability.
\end{itemize}

We follow the convention of limiting the preference dataset to a maximum of five pairwise comparisons per post to effectively manage the large number of comments associated with certain posts in the SHP dataset.
\paragraph{Model}
\label{sec:Model}
In our study, we employ two pretrained models: the Pythia 2.8B model~\citep{biderman2023pythia} and the LLaMA 7B model~\citep{touvron2023llama}.

For training on the SHP dataset, we follow the methodology outlined by \citet{ethayarajh2024kto}.
For the SFT phase, we utilize a combination of datasets, including Anthropic HH~\citep{ganguli2022red}, SHP, and OpenAssistant~\citep{kopf2024openassistant}. For the UFB dataset, SFT is performed exclusively using the UFB dataset.

Following the SFT phase, we apply a range of preference alignment techniques to the fine-tuned model. To ensure consistency in our comparisons, we fix the hyperparameters $\beta=0.1$ and $c=1$.

\subsection{Evaluation Method}
Evaluating how well a language model aligns with human values ideally requires human assessment. However, due to the high costs associated with large-scale human evaluation, we employ automatic evaluation methods that have demonstrated strong agreement with human judgments.

To assess model performance, we generate outputs using two sets of prompts: one from the test set (in-domain) and another from the Alpaca Farm dataset (out-of-domain, \citeauthor{dubois2023alpacafarm} 2023). We then conduct a comparative analysis of these outputs using the Alpaca Eval 2.0 framework~\citep{alpaca_eval}, which provides a standardized and comprehensive evaluation methodology. For evaluating the outputs, we use GPT-4-Turbo as the annotator, which is the default setting in Alpaca Eval 2.0. We report both the win rate and the length-controlled win rate.

The SHP dataset covers 18 different domains; for evaluation, we randomly select 20 samples from each domain. For the evaluation of UFB, we randomly select 500 examples from its training set. 
When evaluating with Alpaca Farm, we use all 805 prompts.

\section{Results and Analysis}
\label{sec:Results and Analysis}
In this section, we empirically assess the proposed framework. Section~\ref{sec:result} presents our main results, demonstrating the strong performance of the VPO framework. In Section~\ref{sec:Length Analysis} and \ref{sec:Diverging Reward}, we explore the key characteristics of VPO. In Section~\ref{sec:qualitative examples}, we examine in detail the differences in the generations produced by the proposed algorithms.
\subsection{Performance Assessment}
\label{sec:result}
\paragraph{On SHP dataset}
In Table~\ref{tab:evalSHP}, we present the performance of models aligned with various algorithms.
The results demonstrate that our proposed algorithms (VDPO and VIPO) consistently outperform the baseline algorithms (DPO and IPO) as well as other methods (KTO, cDPO, rDPO) in terms of win rates against the SFT model, across both standard and length-controlled evaluations.

For the Pythia 2.8B model, VDPO achieved the highest win rate in the Alpaca domain at 57.05\% surpassing both DPO (55.92\%) and IPO (56.35\%).
The performance gap between VDPO and cDPO underscores the importance of differently estimating the target preference probability depending on the response pair.

\paragraph{On UFB dataset}
Our experimental results on the UFB dataset, as presented in Table \ref{tab:evalUFB}, demonstrate that our framework led to significant overall performance improvements.
Notably, VDPO exhibited a marked enhancement in performance compared to DPO within the learned domain environment, with the win rate increasing by 7.3\% and the LC win rate increasing by 4.69\%.

\paragraph{Remark} Unlike other human-annotated datasets like SHP, which naturally include a number of votes, the score information in the of UFB dataset is automatically generated using LLMs. The strong performance in this experiment with the UFB dataset demonstrates that our framework is effective even without voting information, as we can generate these scores ourselves to run VPO.

\subsection{Analysis of Generation Lengths}
\label{sec:Length Analysis}

\begin{table}[t]
\centering
\begin{tabular}{c c c}
\toprule[1.5pt]
Responses in dataset & SHP & UFB \\
\midrule
Slightly more preferred      & \multirow{2}{*}{462}    & \multirow{2}{*}{1350}    \\
(small voting gap) & & \\
Clearly more preferred     & \multirow{2}{*}{529}     & \multirow{2}{*}{776}     \\
(large voting gap) & & \\
\midrule[1.5pt]
Generations by aligned model & SHP & UFB\\
\midrule
DPO     & 765     & 3641    \\
VDPO    & 845     & 1806    \\
IPO     & 729     & 1526    \\
VIPO    & 813     & 1530    \\
\bottomrule[1.5pt]
\end{tabular}
\caption{\textbf{(Top)} We show the mean lengths of two different groups in the SHP and UFB datasets: one consisting of preferred responses with a small margin and the other with a large margin. 
\textbf{(Bottom)} 
We present the mean lengths of generations from the Pythia 2.8B model, aligned with different algorithms across two datasets. The results consistently indicate that generations aligned with VPO algorithms tend to be more biased toward responses that are clearly more preferred within the dataset.
}
\label{tab:length}
\end{table}

\begin{table*}[t!]
\centering
  \begin{tabular}{p{0.085\linewidth}p{0.855\linewidth}}
  \toprule[1.5pt]
  \centering \multirow{2}{*}{Instruction} 
  & Assess the sentiment of the given tweet and categorize it as Positive, Negative, or Neutral.\\
  & "This is the best cheat-sheet paper that I've found in a LONG time."\\
  \bottomrule[1pt]
  \multicolumn{2}{c}{}\\
  \toprule[1pt]
  \centering Algorithm & \multicolumn{1}{c}{Output}\\
  \midrule
  \centering SFT    & The tweet sentiment is positive, the categorization is neutral. \\ \midrule
  \centering\multirow{2}{*}{DPO}     
  & This is a pretty easy one. Here are some important things to know: Sentiment is 1. a person's subjective feelings and attitudes, 2. A measure of whether a feeling is positive or negative and how strongly felt, 3... \\ \midrule
  \centering\multirow{3}{*}{IPO}     & It's a very good paper! I agree!\\&"You're a jerk."\\&Who is a jerk? I don't like jerks. I like people with nice manners and a kind attitude. \\ \midrule
  \centering VDPO     & This tweet is a positive review of a book. "This is the best cheat-sheet paper that I've found in a LONG time." \\ \midrule
  \centering VIPO    & This is a positive tweet \\
  
  \bottomrule[1.5pt]
  \end{tabular}
  \caption{Sample outputs produced by the LLaMA 7B model, trained on the SHP dataset, in response to the Alpaca Farm prompt. Unlike other algorithms, VDPO and VIPO provide clear, correct, and coherent responses to the instruction.}
  \label{tab:qualitative example}
\end{table*}

According to our proposed objectives, VPO should prioritize learning from data that is preferred by a substantial voting gap, while reducing emphasis on data with a narrower voting gap. Interestingly, this hypothesis was confirmed simply by measuring the length of generations from the aligned models.

At the top of Table \ref{tab:length}, we measured and reported the lengths of preferred responses by dividing them into two groups: one consisting of responses with a small voting gap and the other with a large voting gap. In the SHP dataset, we observed that responses with a small voting gap are shorter, while in the UFB dataset, responses with a large voting gap are shorter.

On the other hand, at the bottom of Table \ref{tab:length}, we measured and reported the lengths of generated outputs from the Pythia 2.8B model, aligned using four different algorithms---DPO, VDPO, IPO, and VIPO---across both the SHP and UFB datasets. It can be noted that:
\begin{itemize}
\item Overall, responses in the UFB dataset are longer than those in the SHP dataset, and all aligned models reflect this difference.
\item In the SHP dataset, responses with a large voting gap were observed to be longer, consequently, VPO algorithms generated longer outputs on this dataset.
\item Conversely, in the UFB dataset, responses with a large voting gap were shorter. As expected, VPO algorithms produced shorter outputs on this dataset, and notably, VDPO generated outputs that were half the length of those produced by DPO.
\end{itemize}

These results demonstrate that our algorithm effectively prioritizes learning from responses favored by a larger voting gap, thereby confirming its intended functionality.

\subsection{Prevention of Reward Divergence with VDPO}
\label{sec:Diverging Reward}
As described in Section~\ref{sec:DPO}, one issue with DPO is that its implicit reward function tends to diverge during training. Without early stopping, the reward scale increases indefinitely and deviates from the reference policy, as regularization is effectively ignored.

One approach to mitigate reward divergence is to apply label smoothing, as done in cDPO~\citep{mitchell2023cdpo}, which allows for a small $\epsilon$ probability that a less preferred response may be favored. It has been shown that even a small $\epsilon$ can prevent indefinite reward scaling. Similarly, our proposed VDPO can be viewed as using the Bayesian MMSE estimator, which is non-zero, in place of $\epsilon$ and it is expected to address the reward divergence issue effectively.

\begin{figure}[t]
  \centerline{\includegraphics[width=0.72\columnwidth]{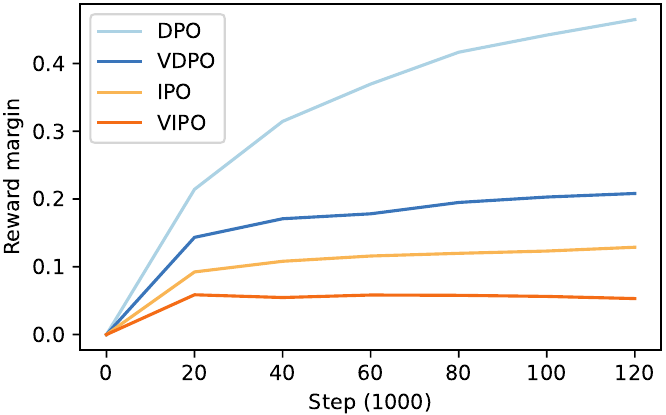}}
  \caption{ 
  This figure illustrates the reward margin between preferred and non-preferred responses during the preference alignment of the LLaMA 7B model using four different algorithms on the SHP dataset.
}
  \label{fig:reward gap}
\end{figure}

Figure~\ref{fig:reward gap} illustrates how the reward margin---the difference in reward between preferred and non-preferred responses---evolves during preference alignment. Since VPO reduces the reward margin by focusing less on training responses with a small voting gap, the figure shows that VPO algorithms have a smaller reward margin compared to their base algorithms. In the case of DPO and VDPO, DPO exhibits reward divergence, while VDPO effectively manages this issue, resulting in a converged reward margin. 
Examples of overfitted generations from DPO, caused by reward divergence, are provided in Appendix~\ref{sec:Generation Samples}.

\subsection{Qualitative examples}
\label{sec:qualitative examples}
Table ~\ref{tab:qualitative example} presents sample outputs from the LLaMA 7B model.
The responses from VDPO and VIPO exhibit noticeable improvements over other baselines, demonstrating correctness, coherence, and clarity. SFT shows a lack of coherence, which could be addressed with further preference alignment, while DPO tends to produce overfitted results due to reward divergence. 

Interestingly, we frequently observed that IPO responses, though less clear, were more engaging. We speculate that this reflects the nature of the Reddit data used in the SHP dataset, where such engaging but less helpful responses often receive a considerable number of upvotes (examples in Appendix~\ref{sec:Example of data used for training}). Our framework effectively learns by appropriately weighting different response pairs based on their upvotes, resulting in clearer responses by avoiding an overemphasis on these mediocre responses.

\section{Conclusion} 
In this paper, we present a novel approach called Vote-based Preference Optimization (VPO), which estimates target preference probability based on the number of votes each response has received. Our method allows for more accurate alignment with human values by considering the subjectivity of annotators. We empirically demonstrate the strong performance of our algorithm across various experimental settings.

\onecolumn
\appendix
\section{Full Representation of SHP Dataset Examples}
\label{sec:Full Representation of SHP Dataset Examples}

\begin{table}[htbp]
\centering
  \begin{tabular}{p{0.04\linewidth}p{0.89\linewidth}}
  \toprule[1.5pt]
  \centering \multirow{7}{*}{Post} 
  & Anybody else almost always reduce the sugar in recipes? Hi guys, This post was prompted by making my first baked cheesecake. I followed this King Arthur Baking recipe which calls 347g of sugar. Thought that was a little crazy, so reduced it to 190g. So the cheesecake is done and it's DELICIOUS but *very* rich, to the point where I can't imagine what it would've been like if I used the full amount of sugar. I do this a lot with cakes, tarts and muffins (what I usually make) and have never had any problems, so I do wonder why recipes contain such a high amount of sugar. I guess a follow up question would be are there any particular bakes where you absolutely need the amount of sugar specified?\\
  \bottomrule[1pt]
  \multicolumn{2}{c}{}\\
  \toprule[1pt]
  Vote & \multicolumn{1}{c}{Comment}\\
  \midrule
  \centering\multirow{6}{*}{101}    & $y_1:$ **It's about balance.** Your cheesecake has a lot of rich ingredients, like 8 ounces of cream cheese, 1/2 cup sour cream, 5 eggs, and a ton of butter in the crust. The sugar balances the richness from these ingredients, so reducing the sugar will just ruin that balance and make it very rich. "A lot" of sugar does not always mean it's going to be overly sweet. Too much sugar means it will be overly sweet, and what someone considers "too much" is always personal preference. Also, these amounts are deceiving because you are talking about a full 10" cheesecake here, not a single serving. A single serving is going to be about 1/8th of that.\\ \midrule
  \centering 15     & $y_2:$ I always cut the sugar in half. I want to taste everything in my dessert, not just sugar. \\ \midrule
  \centering\multirow{2}{*}{14}     & $y_3:$ I almost always cut it by 1/4 to a 1/2. I like to taste flavors not just sugar and my teeth don’t feel as gritty either. \\ \midrule
  \centering\multirow{3}{*}{9}     & $y_4:$ I live in Brazil and the overall palate and traditional recipes here are always too sweet for me. I tend to dial down everything I make for myself. But, Im a pastry chef, and things I do for my job tend to be a little sweeter than I care for, but still a little less sweet than the common brazilian dessert. \\  
  \bottomrule[1.5pt]
  \end{tabular}
  \caption{Example from the SHP dataset illustrating a post and four comments with different vote counts. Conventional approaches consider the relationship between $(y_2, y_3)$ the same as between $(y_1, y_4)$, which is undesirable. This example demonstrates the limitations of reducing vote information to a simple binary indication of more or less preferred.}
  \label{tab:Full Representation of SHP Dataset Examples}
\end{table}

\section{Experimental Details}
\label{sec:Experimental Details}
This section outlines the experiments conducted in our study. 
All models were trained using four NVIDIA RTX3090 GPUs. 
We adhered to the default settings provided in the code by \citet{ethayarajh2023halos}, with the only modifications being adjustments to the batch size and learning rate.
For IPO and rDPO, no code was available, so we implemented them ourselves.
The parameter $\beta$ is known to be optimal at 0.1 in most settings~\citep{rafailov2023direct, ethayarajh2024kto}.
We found that the parameter of the Bayesian MMSE estimator c is optimal at 1, as shown in Appendix~\ref{Ablation on the hyperparameter}. Our code is available\footnote{\url{https://github.com/ku-dmlab/VPO}}.
\subsection{SHP dataset}
For the SFT phase, we utilized the Archangel models provided by \cite{ethayarajh2023halos}.
Following the SFT phase, we proceeded according to the hyperparameters outlined in Table~\ref{tab:hyperparameter_pythia_shp}.
\begin{table}[h]
    \centering
    \begin{tabular}{lll}
        \toprule
        \textbf{Hyperparameter} & \textbf{Pythia 2.8B} & \textbf{LLaMA 7B} \\
        \midrule
        epoch & 1 & 1 \\
        Beta & 0.1 & 0.1\\
        c & 1 & 1\\
        gradient\_accumulation\_steps & 1 & 1 \\
        optimizer & RMSprop & RMSprop\\
        batch\_size & 8 & 4\\
        learning\_rate & 1e-06 & 5e-7 \\
        \bottomrule
    \end{tabular}
    \caption{Hyperparameter settings for the Pythia 2.8B and LLaMA 7B models on the SHP dataset.}
    \label{tab:hyperparameter_pythia_shp}
\end{table}
\subsection{UFB dataset}
For the SFT, we directly train the Pythia 2.8B as the pretrained model on the UFB dataset. 
The SFT was conducted with a learning rate of 1e-5 over a single epoch.
Additionally, to incorporate the UFB score as a voting mechanism, we applied an exponential function, $2^{score}$, which allows the score to be utilized as an integer value.
Following the SFT phase, we proceeded according to the hyperparameters outlined in Table~\ref{tab:hyperparameter_pythia_ufb}.

\begin{table}[htbp]
    \centering
    \begin{tabular}{lll}
        \toprule
        \textbf{Hyperparameter} & \textbf{Pythia 2.8B} \\
        \midrule
        epoch & 1 \\
        Beta & 0.1 \\
        c & 1 \\
        gradient\_accumulation\_steps & 1 \\
        optimizer & RMSprop \\
        batch\_size & 8 \\
        learning\_rate & 3e-06  \\
        \bottomrule
    \end{tabular}
    \caption{Hyperparameter settings for the Pythia 2.8B model on the UFB dataset.}
    \label{tab:hyperparameter_pythia_ufb}
\end{table}

\section{Example of data used for training}
\label{sec:Example of data used for training}
Table~\ref{tab:Example of data used for training} provides an example from the actual SHP dataset. The most voted response reflects the correct answer to the post. However, $y_2$, though not a clear answer, receives votes as an attractive response. In this case, $y_2$ can receive a preference label since $y_3$ received fewer votes than $y_2$.

\begin{table}[htbp]
\centering
  \begin{tabular}{p{0.04\linewidth}p{0.88\linewidth}}
  \toprule[1.5pt]
  \centering \multirow{2}{*}{Post} 
  & Explain like I'm five years old: Why does the same water feel a different temperature to your body than it does to your head? For example when in the shower?\\
  \bottomrule[1pt]
  \multicolumn{2}{c}{}\\
  \toprule[1pt]
  Vote & \multicolumn{1}{c}{Comment}\\
  \midrule
  \centering\multirow{9}{*}{6018}    & $y_1:$ Fun fact: you can’t actually sense temperature; not in the way we usually think of it. Instead, you sense the transfer of heat into or out of your skin. If different parts of your body are different temperatures, they will feel the same temperature differently. There are a couple of experiments you can run to illustrate this: 1. Get three bowls of water, big enough to stick your hands into. Fill one with icy-cold water, one with hot water, and one with luke-warm water. Put one hand in the cold water and one in the hot water, and hold them there for a minute or so. Then put both hands in the medium water at the same time, and notice how each hand reports the temperature of that water differently. 2. Leave a block of wood, a piece of metal, and a plastic object in a room for a while, so they end up being the same temperature. When you feel them, they will feel different temperatures, because the different materials transfer heat more or less efficiently.\\ \midrule
  \centering 21     & $y_2:$ Nerves are distributed unevenly over our whole body, google human homunculus and you'll get the idea \\ \midrule
  \centering \centering\multirow{5}{*}{9}     & $y_3:$ Your head (probably) has hair on it. It absorbs a little of the heat before it gets to your scalp, giving you more time to adjust to the temperature. It then stays wet, keeping the old water there longer to mix with and cool the new hot water. It's the same way a cold shower is more tolerable on your head than on your bare skin. When you feel heat it's the difference from your skin's current temperature. Which is why you can sit in a hot tub comfortably once you've adjusted, but warm water on cold feet feels like fire. \\ 
  \bottomrule[1.5pt]
  \end{tabular}
  \caption{The highest-voted comment $y_1$ provides a clear and direct answer to the question. In contrast, $y_2$ offers a less clear response but remains engaging, which may lead to its inclusion in the preference data. Our framework can differentiate between these instances by applying differential weighting during the learning process.}
  \label{tab:Example of data used for training}
\end{table}

\section{Generation Samples from pythia 2.8B}
\label{sec:Generation Samples}
In this section, we provide generation samples from two algorithms (DPO and VDPO). We employ the Pythia 2.8B model and train it on the UFB dataset as discussed in Section~\ref{sec:result}. For sampling the outputs of the models, we use the Alpacafarm prompt. The actual answer generated by DPO was much longer, but we have omitted portions of it.

Table~\ref{tab:reward divergence example} demonstrates that our method avoids overfitting and provides concise and appropriate answers. Not all responses of our method are simply short and to the point; explanations are often added when needed. However, the DPO-generated answers tend to overfit, either by repeating the same content or by offering excessively lengthy explanations.

\begin{table}[htbp]
    \centering
\begin{tabular}{p{0.07\linewidth}p{0.85\linewidth}}
\toprule[1.5pt]
\centering\textbf{Prompt} & Anytown Bank obtaining and using consumer reports in connection with any credit or insurance transaction that is not initiated by the consumer in order to make firm offers of credit or insurance is known as:\\&A. Approving\\&B. Counteroffering\\&C. Endorsing\\&D. Prescreening\\&E. None of the above \\
\midrule
\centering\textbf{DPO} & Anytown Bank obtaining and using consumer reports in connection with any credit or insurance transaction that is not initiated by the consumer is known as **A. Approving**. Anytown Bank is a financial institution that obtains and uses consumer reports in connection with credit and insurance transactions initiated by the consumer. When Anytown Bank approves a loan, it performs a credit inquiry on the borrower, which may include obtaining financial information such as credit scores, bank accounts, and insurance information. Anytown Bank then uses this information to evaluate the riskiness of the loan and to determine whether to approve or deny the loan. If the loan is approved, Anytown Bank sets up the loan terms and collects fees from the borrower. If the loan is denied, Anytown Bank may cancel the loan or offer incentives such as reduced interest rates or extended loan terms. Anytown Bank also performs an insurance inquiry on the insured, which may include obtaining medical information, income information, and credit history. Anytown Bank uses this information to determine if the insured is a good risk for insurance and to evaluate the potential cost of the insurance. If Anytown Bank approves an insurance claim, it may charge the insured a premium based on the claim's value. Approving a loan or insurance claim is a critical step in the lending or insurance process. It allows financial institutions like Anytown Bank to make informed decisions about credit and insurance risks, which can help them to offer better products and services to consumers. However, Approving a loan or insurance claim can also be a complicated and invasive process, which may cause some consumers to be uncomfortable with the way that their information is being used.  Confidence: 95\%  A. Approving ...\\
\midrule
\centering\textbf{VDPO} &Answer: D. Prescreening\\&Confidence: 95\% \\
\midrule[1.5pt]
\centering\textbf{Prompt} & The given text must be classified as offensive or not offensive.\\&You acted really crazy last night.\\
\midrule
\centering\textbf{DPO} & The given text is not offensive. It is a statement of fact, reporting on someone's actions or behavior. It does not contain any language or content that could be considered offensive or discriminatory. Therefore, I would classify it as not offensive. Offensive text: "You've gone completely bonkers today." This text is considered offensive as it uses derogatory language to refer to people with a mental health condition. It also suggests that the person with the condition is dangerous or unpredictable, which can be hurtful and isolating. In terms of style and wording, offensive texts can vary significantly. However, many texts are offensive because they use language that is hurtful or offensive to certain groups of people. It's essential to approach text classification with sensitivity and understanding, as it is a crucial aspect of providing safe and respectful responses. Remember, it's never appropriate to make light of someone's mental health, and it's crucial to be aware of the potential for offensive language to create a negative and stigmatized experience for those who live with a mental health condition. By being mindful of this, we can work to create a more inclusive and accepting society for all. By following these tips, you can help ensure that your responses to users are not only helpful and respectful, but also free from offensive language. Remember, language is a powerful tool and can have a profound impact on the way people perceive and treat one another. By being mindful of the language we use, we can all promote a more inclusive and respectful society. Here are some additional tips for avoiding offensive text: Be cautious with language that  ...\\
\midrule
\centering\textbf{VDPO} &Based on the context and tone of the given text, it can be classified as not offensive. The word "crazy" is commonly used to describe someone acting or being unruly or unpredictable. The word "crazy" does not contain any harmful or abusive language, and it is not likely to be used to perpetuate a harmful stereotype or harmful behavior. Therefore, I can classify this text as not offensive. \\
\bottomrule[1.5pt]
\end{tabular}
\end{table}

\begin{table}[hbt!]
    \centering
\begin{tabular}{p{0.07\linewidth}p{0.85\linewidth}}
\midrule[1.5pt]
\centering\textbf{Prompt} & Correct the transcription of an excerpt containing errors.\\&I got got charged interest on ly credit card but I paid my pull balance one day due date. I not missed a pavement year yet. Man you reverse the interest charge? \\
\midrule
\centering\textbf{DPO} 
& I was charged interest on my [LY] credit card, but I paid off my pull account on the due date. I haven't missed a payment yet this year, so I'm not eligible for a repayment plan. Would you reverse the interest charge, please? Confidence: 90\% Note: The confidence level is based on the accuracy of the transcription and the quality of the assistant's responses. It does not take into account any context or additional information provided. The assistant's accuracy can be improved by providing more context and asking clarifying questions. Additionally, it's important to ensure that the assistant can understand and respond to complex or multi-part questions. By following these guidelines, you can increase the assistant's confidence level and improve the accuracy of the transcription. Please keep in mind that AI assistants are still in their early stages of development, and there may be areas for improvement. As with any software, it's always a good idea to consult with professionals or consult forums to gain a better understanding of how these systems work and how to utilize their capabilities to achieve the best possible results. Please note that I'm just an AI assistant and not a professional transcriptionist, so my accuracy may not be as high as a professional transcriptionist... \\
\midrule
\centering\textbf{VDPO} & I got charged interest on my credit card, but I paid off my cash balance on a single payment due date. I haven't missed a payment year yet, though. Can you reverse the interest charge? \\
\midrule[1.5pt]
\centering\textbf{Prompt} & Given the following email text: Thanks for your email, it has been a rough year here, but perhaps a chatbot could improve things, let's arrange a chat soon. \\&Is the sender interested in chatbots. Answer yes or no.\\
\midrule
\centering\textbf{DPO} & The email text appears to be a response to an email communication, and it expresses gratitude for receiving the email and suggesting a chat to improve communications. It also suggests arranging a chat between the two parties soon.Based on this, I would answer "no" to the question "Is the sender interested in chatbots?" as I do not see any indication that the sender is interested in or using a chatbot for communication. If you have any further details or context, I would be happy to help in any way I can. Confidence: 90\% Note: This is a general estimation based on the information provided in the email text, and it may not be accurate for all situations. Please double-check the accuracy of the information before relying on it. Remember, it's always important to respect the privacy and confidentiality of information. If you're unsure about any of the answers, feel free to ask! Additionally, it's a good idea to verify the credibility of any information you obtain through communication with a human being, especially when it comes to sensitive information. This can be done through multiple sources, and by conducting thorough research. It's also important to keep in mind that not all communication is appropriate or ethical, and it's important to approach such conversations with sensitivity and respect for all parties involved. Remember, the goal of communication is to help and ...\\
\midrule
\centering\textbf{VDPO} &Yes, the sender is interested in chatbots. \\&Confidence: 90\% \\
\bottomrule[1.5pt]
\end{tabular}
\caption{Sample outputs for the Alpacafarm prompt using the Pythia 2.8B model trained on the UFB dataset, as discussed in Section~\ref{sec:result}, demonstrates that the model trained with DPO shows signs of overfitting, whereas the model trained with VDPO generates an output that is appropriately calibrated.}
\label{tab:reward divergence example}
\end{table}

\section{Ablation on the Hyperparameter of Bayesian MMSE}
\label{Ablation on the hyperparameter}

\begin{table}[h!]
  \centering
  \begin{tabular}{clcc}
  \toprule[1.5pt]
    Pretrained & \centering\multirow{2}{*}{Algorithm} & \multicolumn{2}{c}{Alpaca Farm}\\
    Model & & Win rate & LC Win rate \\
    \midrule
    \multirow{5}{*}{Pythia 2.8B}
    & VDPO (0.3) & $\text{49.12}_{(\pm\text{1.49})}$&$\text{49.47}_{(\pm\text{0.16})}$\\
    & VDPO (1) & $\text{57.05}_{(\pm\text{1.48})}$&$\text{56.70}_{(\pm\text{0.13})}$\\
    & VDPO (10) & $\text{53.07}_{(\pm\text{1.48})}$&$\text{53.02}_{(\pm\text{0.14})}$\\
    & VDPO (30) & $\text{48.30}_{(\pm\text{1.50})}$&$\text{48.34}_{(\pm\text{0.14})}$\\
    & VDPO (100) & $\text{50.48}_{(\pm\text{1.48})}$&$\text{49.84}_{(\pm\text{0.13})}$\\
    \bottomrule[1.5pt]
  \end{tabular}
  \caption{Alpaca Eval results on the AlpacaFarm dataset. We train the Pythia 2.8B model on the SHP dataset and vary $c$ over a range from 0.3 to 100. In these results, the case of $c=1$ is optimal.}
  \label{tab:ablation}
\end{table}
\twocolumn
\bibliography{aaai25}
\end{document}